\documentclass[10pt,twocolumn,letterpaper]{article}

\usepackage[pagenumbers]{cvpr} 

\usepackage{capt-of}
\usepackage{booktabs}
\usepackage{pifont}
\usepackage{graphicx}
\usepackage{multirow}

\usepackage[table]{xcolor}
\usepackage{pifont}

\definecolor{tickgreen}{HTML}{287D3C}
\definecolor{crossred}{HTML}{B33A3A}
\definecolor{datasetcategory}{HTML}{FFFFFF}
\definecolor{datasetours}{HTML}{FFFFFF}

\newcommand{\cmark}{\textcolor{tickgreen}{\ding{51}}}
\newcommand{\xmark}{\textcolor{crossred}{\ding{55}}}
\newcommand{\pmark}{(\ding{51})}
\newcommand{\na}{--}

\definecolor{cvprblue}{rgb}{0.21,0.49,0.74}
\usepackage[pagebackref,breaklinks,colorlinks,allcolors=cvprblue]{hyperref}

\usepackage{booktabs,graphicx}
\usepackage{tabularx,array}

\def\paperID{*****} 
\def\confName{CVPR}
\def\confYear{2027}

\title{RoboMoCap: A Large-Scale Whole-Body Motion Capture Dataset with Egocentric Vision for Robot Learning}

\author{
Mayur Deshmukh$^{1}$ \quad
Aaron Tung$^{2}$ \quad
Junyi Li$^{2}$ \quad
Nikash Bhardwaj$^{3}$ \\[0.4em]
Ademi Adeniji$^{3}$ \quad
Vincent Liu$^{3}$ \quad
Michael Cho$^{2}$ \quad
Jiafei Duan$^{1}$ \\[0.8em]
$^{1}$MAGIC Lab, National University of Singapore \quad
$^{2}$BitRobot \quad
$^{3}$Grounded Superintelligence \\[0.8em]
\url{https://bitrobot-foundation.github.io/robomocap/}
}

\begin{document}

\newcommand{\egocomparison}{
\begin{table*}[t]
    \centering
    \caption{\textbf{Egocentric full-body motion reconstruction on RoboMoCap.}
Comparison of existing methods across joint position and rotation accuracy, temporal motion consistency, and head pose estimation.}
    \label{tab:egomocap_reconstruction}

    \setlength{\tabcolsep}{3pt}
    \renewcommand{\arraystretch}{1.15}

    \resizebox{\textwidth}{!}{%
    \begin{tabular}{lcccccccccc}
        \toprule
        & \multicolumn{4}{c}{\textbf{Joint Position}}
        & \textbf{Rotation}
        & \multicolumn{3}{c}{\textbf{Temporal Motion}}
        & \multicolumn{2}{c}{\textbf{Head}} \\
        \cmidrule(lr){2-5}
        \cmidrule(lr){6-6}
        \cmidrule(lr){7-9}
        \cmidrule(lr){10-11}

        \textbf{Method}
        & \shortstack{MPJPE\\(mm) $\downarrow$}
        & \shortstack{PA-MPJPE\\(mm) $\downarrow$}
        & \shortstack{W-MPJPE\\(mm) $\downarrow$}
        & \shortstack{WA-MPJPE\\(mm) $\downarrow$}
        & \shortstack{MPJRE\\($^\circ$) $\downarrow$}
        & \shortstack{MPJVE\\(cm/s) $\downarrow$}
        & \shortstack{Accel. Error\\(cm/s$^2$) $\downarrow$}
        & \shortstack{Jitter\\(m/s$^3$)$\downarrow$}
        & \shortstack{Position\\(mm) $\downarrow$}
        & \shortstack{Orientation\\($^\circ$) $\downarrow$} \\
        \midrule

        AGRoL~\cite{du2023avatars}
        & 82.3 & \textbf{51.2} & \textbf{95.4} & \textbf{64.9} & \textbf{28.2} & 15.1 & 285.7 & 119.31 & \textbf{30.4} & 23.0 \\

        EgoPoser~\cite{jiang2024egoposer}
        & 88.8 & 56.1 & 105.8 & 71.4 & \textbf{28.2} & 17.1 & 251.5 & 96.48 & \textbf{30.4} & 23.0 \\

        EgoEgo~\cite{li2023ego}
        & 130.9 & 97.2 & 224.6 & 118.8 & 33.5 & 27.5 & 426.7 & 157.73 & 138.2 & \textbf{22.9} \\

        EgoAllo~\cite{yi2025estimating}
        & \textbf{80.8} & 69.2 & 103.9  & 79.3 & 30.3 & \textbf{14.5} & \textbf{214.5} & \textbf{72.06} & 74.8 & \textbf{22.9} \\
        

        \bottomrule
    \end{tabular}%
    }
\end{table*}
}

\newcommand{\lowerbodycomparison}{%
\begin{table}[t]
\centering
\caption{\textbf{Foot contact, placement, and physical plausibility on RoboMoCap.}
Contact predictions are derived from reconstructed foot joints using a shared height- and velocity-based rule, calibrated on training participants and evaluated against sole-marker ground truth.
Footprint and Step measure planted-foot position and step displacement errors, respectively; Skate measures foot velocity during ground-truth contact, and Penetr. measures ground penetration.
GT SMPL-X is included as a reference.}
\small
\setlength{\tabcolsep}{3pt}
\resizebox{\columnwidth}{!}{%
\begin{tabular}{lccccc}
\toprule
 & Contact & \multicolumn{2}{c}{Foot placement} & \multicolumn{2}{c}{Plausibility} \\
\cmidrule(lr){2-2}\cmidrule(lr){3-4}\cmidrule(lr){5-6}
Method
 & \shortstack{F1 (\%)\\$\uparrow$}
 & \shortstack{Footprint\\(mm) $\downarrow$}
 & \shortstack{Step\\(mm) $\downarrow$}
 & \shortstack{Skate\\(mm/s) $\downarrow$}
 & \shortstack{Penetr.\\(mm) $\downarrow$} \\
\midrule
EgoEgo~\cite{li2023ego}           & 74.3 & 220.1 & 234.7 & 291.1 & 36.0 \\
EgoAllo~\cite{yi2025estimating}   & \textbf{90.5} & \textbf{85.1} & \textbf{154.1} & \textbf{54.8} & \textbf{8.5} \\
EgoPoser~\cite{jiang2024egoposer} & 68.1 & 140.3 & 266.3 & 314.4 & 64.1 \\
AGRoL~\cite{du2023avatars}        & 75.3 & 113.8 & 237.0 & 273.7 & 66.8 \\
\midrule
\textit{GT SMPL-X} & \textit{93.6} & -- & -- & \textit{43.6} & \textit{0.0} \\
\bottomrule
\end{tabular}%
}
\label{tab:feet}
\end{table}
}

\newcommand{\kinematicfidelity}{
\begin{table*}[t]
\centering
\caption{\textbf{Human-to-humanoid retargeting: kinematic fidelity on RoboMoCap.}
Comparison of human-to-Unitree G1 motion retargeting methods across 543 test sessions ($\approx$18\,h), evaluating global and root-relative positional errors for matched key bodies, feet, and hands/wrists, alongside orientation accuracy.}
\label{tab:egomocap_retarget_kinematic}
\small
\setlength{\tabcolsep}{5pt}
\renewcommand{\arraystretch}{1.15}
\begin{tabular}{lccccccc}
\toprule
 & \multicolumn{2}{c}{\textbf{Matched Key Bodies}}
 & \multicolumn{2}{c}{\textbf{Feet}}
 & \multicolumn{2}{c}{\textbf{Hands/Wrists}}
 & \textbf{Orientation} \\
\cmidrule(lr){2-3}\cmidrule(lr){4-5}\cmidrule(lr){6-7}
\cmidrule(lr){8-8}
\textbf{Method}
 & \shortstack{$E_{\mathrm{g\text{-}mpbpe}}$\\(mm) $\downarrow$}
 & \shortstack{$E_{\mathrm{mpbpe}}$\\(mm) $\downarrow$}
 & \shortstack{Global\\(mm) $\downarrow$}
 & \shortstack{Root-rel.\\(mm) $\downarrow$}
 & \shortstack{Global\\(mm) $\downarrow$}
 & \shortstack{Root-rel.\\(mm) $\downarrow$}
 & \shortstack{Geodesic\\(rad) $\downarrow$} \\
\midrule
GMR \cite{araujo2026retargeting}         & 90.7 & 78.6 & 84.8 & 87.7 & 131.5 & 121.7 & \textbf{0.385} \\
PHC \cite{luo2023perpetual}              & 173.9 & 64.9 & 160.6 & 62.7 & 186.1 & 88.8 & 0.435  \\
OmniRetarget \cite{yang2026omniretarget} & \textbf{34.8} & \textbf{36.3} & \textbf{49.5} & \textbf{61.8} & \textbf{29.3} & \textbf{28.8} & 0.470 \\
\bottomrule
\end{tabular}
\end{table*}
}



 
\newcommand{\kinematicfeasibility}{
\begin{table*}[t]
\centering
\caption{\textbf{Retargeting feasibility and contact preservation on RoboMoCap.}
Comparison of human-to-Unitree G1 motion retargeting methods across 543 test sessions ($\approx$18\,h), evaluating foot skating, ground penetration, floating, contact preservation, joint-limit and self-collision violations, and clip-level success.}
\label{tab:egomocap_retarget_feasibility}
\setlength{\tabcolsep}{3pt}
\renewcommand{\arraystretch}{1.15}
\resizebox{\textwidth}{!}{%
\begin{tabular}{lccccccccc}
\toprule
 & \multicolumn{2}{c}{\textbf{Foot Skating}}
 & \multicolumn{2}{c}{\textbf{Ground Penetration}}
 & \textbf{Floating}
 & \textbf{Contact F1}
 & \multicolumn{2}{c}{\textbf{Constraints}}
 & \textbf{Success} \\
\cmidrule(lr){2-3}\cmidrule(lr){4-5}\cmidrule(lr){6-6}
\cmidrule(lr){7-7}\cmidrule(lr){8-9}\cmidrule(lr){10-10}
\textbf{Method}
 & \shortstack{Rate\\(\%) $\downarrow$}
 & \shortstack{p99 speed\\(m/s) $\downarrow$}
 & \shortstack{Frames\\(\%) $\downarrow$}
 & \shortstack{Max depth\\(cm) $\downarrow$}
 & \shortstack{Mean gap\\(cm) $\downarrow$}
 & \shortstack{Foot\\(\%) $\uparrow$}

 & \shortstack{Joint-limit\\(\%) $\downarrow$}
 & \shortstack{Self-coll.\\(\%) $\downarrow$}
 & \shortstack{Clips\\(\%) $\uparrow$} \\
\midrule
GMR \cite{araujo2026retargeting}         & 12.3 & 0.36 & 72.5 & 6.8 & 1.5 & 83.0 & 87.2 & 10.7 & 7.4 \\
PHC \cite{luo2023perpetual}              & 18.5 & 0.56 & 20.9 & 37.1 & 1.4 & 80.4 & 82.9 & \textbf{1.1} & \textbf{79.6} \\
OmniRetarget \cite{yang2026omniretarget} & \textbf{2.2} & \textbf{0.26} & \textbf{0.0} & \textbf{0.2} & \textbf{0.1} & \textbf{87.7}  & \textbf{71.8} & 10.0 & 53.6 \\
\bottomrule
\end{tabular}%
}
\end{table*}
}

\newcommand{\retargetingcomparison}{%
\begin{table*}[t]
\centering
\caption{\textbf{Human-to-humanoid retargeting on RoboMoCap: kinematic fidelity, feasibility and contact preservation.}}
\label{tab:egomocap_retarget}
\resizebox{\textwidth}{!}{%
\begin{tabular}{l*{16}{c}}
\toprule
 & \multicolumn{7}{c}{\textbf{Kinematic fidelity}}
 & \multicolumn{9}{c}{\textbf{Feasibility and contact preservation}} \\
\cmidrule(lr){2-8}\cmidrule(lr){9-17}
 & \multicolumn{2}{c}{\textbf{Matched Key Bodies}}
 & \multicolumn{2}{c}{\textbf{Feet}}
 & \multicolumn{2}{c}{\textbf{Hands/Wrists}}
 & \textbf{Orientation}
 & \multicolumn{2}{c}{\textbf{Foot Skating}}
 & \multicolumn{2}{c}{\textbf{Ground Penetration}}
 & \textbf{Floating}
 & \textbf{Contact F1}
 & \multicolumn{2}{c}{\textbf{Constraints}}
 & \textbf{Success} \\
\cmidrule(lr){2-3}\cmidrule(lr){4-5}\cmidrule(lr){6-7}
\cmidrule(lr){8-8}\cmidrule(lr){9-10}\cmidrule(lr){11-12}
\cmidrule(lr){13-13}\cmidrule(lr){14-14}
\cmidrule(lr){15-16}\cmidrule(lr){17-17}
\textbf{Method}
 & \shortstack{$E_{\mathrm{g\text{-}mpbpe}}$\\(mm) $\downarrow$}
 & \shortstack{$E_{\mathrm{mpbpe}}$\\(mm) $\downarrow$}
 & \shortstack{Global\\(mm) $\downarrow$}
 & \shortstack{Root-rel.\\(mm) $\downarrow$}
 & \shortstack{Global\\(mm) $\downarrow$}
 & \shortstack{Root-rel.\\(mm) $\downarrow$}
 & \shortstack{Geodesic\\(rad) $\downarrow$}
 & \shortstack{Rate\\(\%) $\downarrow$}
 & \shortstack{p99 speed\\(m/s) $\downarrow$}
 & \shortstack{Frames\\(\%) $\downarrow$}
 & \shortstack{Max depth\\(cm) $\downarrow$}
 & \shortstack{Mean gap\\(cm) $\downarrow$}
 & \shortstack{Foot\\(\%) $\uparrow$}
 & \shortstack{Joint-limit\\(\%) $\downarrow$}
 & \shortstack{Self-coll.\\(\%) $\downarrow$}
 & \shortstack{Clips\\(\%) $\uparrow$} \\
\midrule
GMR~\cite{araujo2026retargeting}
 & 90.7 & 78.6 & 84.8 & 87.7 & 131.5 & 121.7 & \textbf{0.385}
 & 12.3 & 0.36 & 72.5 & 6.8 & 1.5 & 83.0 & 87.2 & 10.7 & 7.4 \\
PHC~\cite{luo2023perpetual}
 & 173.9 & 64.9 & 160.6 & 62.7 & 186.1 & 88.8 & 0.435
 & 18.5 & 0.56 & 20.9 & 37.1 & 1.4 & 80.4 & 82.9 & \textbf{1.1} & \textbf{79.6} \\
OmniRetarget~\cite{yang2026omniretarget}
 & \textbf{34.8} & \textbf{36.3} & \textbf{49.5} & \textbf{61.8}
 & \textbf{29.3} & \textbf{28.8} & 0.470
 & \textbf{2.2} & \textbf{0.26} & \textbf{0.0} & \textbf{0.2}
 & \textbf{0.1} & \textbf{87.7} & \textbf{71.8} & 10.0 & 53.6 \\
\bottomrule
\end{tabular}%
}
\end{table*}
}



\newcommand{\taxonomy}{%
\begin{table*}[t]
\centering
\caption{\textbf{RoboMoCap action taxonomy.}
The dataset defines 291 action protocols across eight categories, spanning locomotion, loco-manipulation, dynamic recovery, collaborative interactions, and long-horizon activities. The dataset comprises over 200 hours of multimodal recordings from 14 participants.}
\label{tab:action_taxonomy}

\footnotesize
\setlength{\tabcolsep}{3pt}
\renewcommand{\arraystretch}{1.17}

\begin{tabularx}{\textwidth}{
@{}
>{\raggedright\arraybackslash}p{3.2cm}
>{\centering\arraybackslash}p{0.65cm}
>{\raggedright\arraybackslash}X
>{\raggedright\arraybackslash}X
@{}
}
\toprule
\textbf{Action category} &
\textbf{\#} &
\textbf{Representative tasks} &
\textbf{Task variations} \\
\midrule

\textbf{Locomotion primitives}
& 44
& Variable-speed walking, turns and figure-eights, backward and
sideways walking, narrow passages, ramps and stairs
& Speed, acceleration, travel direction, path curvature,
stance width, obstacles and terrain \\
\addlinespace[2pt]

\textbf{Locomotion posture and dynamics}
& 60
& Sit-to-stand, squatting, kneeling, lying and rising, jumping,
mimed sports and combat, dance, running and cutting
& Posture, support configuration, movement intensity,
in-place vs.\ travelling, gait transitions \\
\addlinespace[2pt]

\textbf{Loco-manipulation}
& 86
& Height-paired transport, varied-load carries, carrying over
terrain and stairs, manipulation while walking, bin loading,
precision placement, deformable loads
& Object weight and shape, carry position, pick and place height,
route and terrain, concurrent manipulation \\

\textbf{Static manipulation}
& 44
& Stationary pick-and-place, drawer, lid and
latch operation, folding and pouring, tool use, reach-envelope mapping
& Reach height and direction, articulation type,
uni- vs.\ bimanual coordination, object rigidity \\
\addlinespace[2pt]

\textbf{Long-horizon composite}
& 11
& Delivery shelving, order picking, restocking, palletising,
sorting, kit assembly and room tidying
& Task order, object selection, goal specification,
start-state and resource perturbations \\
\addlinespace[2pt]

\textbf{Dynamic and recovery}
& 15
& Recovery stepping, trips, surface changes, reduced friction,
load perturbations, controlled falls and rising
& Perturbation source and direction, support surface,
carried load and recovery strategy \\
\addlinespace[2pt]

\textbf{Human--human loco-manipulation}
& 14
& Static and walking handovers, two-person lifting and carrying,
relay chains, assisted lifts, passing in narrow spaces
& Partner movement, object weight and length, relative position, load sharing and route constraints\\
\addlinespace[2pt]

\textbf{Environment-coupled whole-body}
& 17
& Cart pushing and pulling, cart slalom and parking, sled dragging,
bracing against fixed loads, hoisting and lowering
& Resistance type, load, direction, obstacle layout
and sustained force \\

\midrule
\textbf{Total} & \textbf{291} & & \\
\bottomrule

\end{tabularx}
\end{table*}%
}

\newcommand{\composition}{%
\begin{table}[t]
    \centering
    \caption{\textbf{RoboMoCap release-candidate composition.}
    Hours are nominal session durations, not yet validated
    synchronised multimodal hours. Recorded IDs are metadata
    identifiers, not necessarily distinct semantic action classes.
    Figures are rounded.}
    \label{tab:dataset_composition}
    \small
    \setlength{\tabcolsep}{3.5pt}
    \begin{tabular}{lrrr}
        \toprule
        \textbf{Action family} & \textbf{Sessions} &
        \textbf{IDs} & \textbf{Hours} \\
        \midrule
        Locomotion primitives          & 1,288 & 44 & 44.67 \\
        Posture and dynamics           & 813   & 60 & 17.83    \\
        Loco-manipulation              & 2,465 & 86 & 87.7 \\
        Static manipulation            & 398   & 44 & 15.02    \\
        Environment-coupled whole-body & 304   & 17 & 9.79  \\
        Dynamic recovery               & 154   & 15 & 5.52    \\
        Human--human loco-manipulation & 134   & 14 & 4.58  \\
        Long-horizon composite         & 163   & 11 & 6.65    \\
        \midrule
        \textbf{Total} & \textbf{5,719} &
        \textbf{291} & \textbf{191.22} \\
        \bottomrule
    \end{tabular}
\end{table}
}

\newcommand{\datasetcomparison}{%
\begin{table*}[t]
\centering
\caption{\textbf{Comparison of RoboMoCap with related datasets.} \cmark: provided; \pmark: partial, estimated, limited coverage, or third-party release;
\xmark: absent; \na: not applicable or not reported. Year: publication venue year, or arXiv/submission year otherwise. Body GT: Opt = optical marker mocap, IMU = inertial suit, MV = markerless multi-view fitting,
Ps = pseudo-labels from video or wearable devices, Syn = synthetic.
\#Ego: head-mounted cameras; \#Wrist: wrist-mounted cameras; Body-fac.: at least one egocentric camera observes the wearer's body.
Loco: locomotion beyond forward walking (backward, lateral, turning, speed).
L-M: loco-manipulation, e.g., carrying or transport while walking. Frame counts (fr) are given when hours are not reported.}
\label{tab:dataset_comparison}
\resizebox{\textwidth}{!}{%
\begin{tabular}{l c r r l c c c c c c c c c c c}
\toprule
Dataset & Year & Hours & \#Subj & Body GT & \#Ego & Body-fac. & \#Wrist & \#Exo & IMU & Hand & Obj.\ 6D & Scene & Audio & Loco & L-M \\
\midrule
\rowcolor{datasetcategory}
\multicolumn{16}{l}{\textbf{\textit{Motion capture and pseudo-labelled motion}}} \\
\midrule
AMASS~\cite{AMASS:ICCV:2019} & 2019 & 40+ & 344 & Opt & 0 & \xmark & \xmark & \xmark & \xmark & \pmark & \xmark & \xmark & \xmark & \cmark & \pmark \\
LAFAN1~\cite{harvey2020robust} & 2020 & 4.6 & 5 & Opt & 0 & \xmark & \xmark & \xmark & \xmark & \xmark & \xmark & \xmark & \xmark & \cmark & \xmark \\
GRAB~\cite{taheri2020grab} & 2020 & 1.6M fr & 10 & Opt & 0 & \xmark & \xmark & \xmark & \xmark & \cmark & \cmark & \xmark & \xmark & \xmark & \xmark \\
100STYLE~\cite{mason2022real} & 2022 & 22.1 & 1 & Opt & 0 & \xmark & \xmark & \xmark & \xmark & \xmark & \xmark & \xmark & \xmark & \cmark & \xmark \\
OMOMO~\cite{li2023object} & 2023 & 10 & 17 & Opt & 0 & \xmark & \xmark & \xmark & \xmark & \xmark & \cmark & \xmark & \xmark & \pmark & \cmark \\
Motion-X~\cite{lin2023motion} & 2023 & 144.2 & \na & Ps & 0 & \xmark & \xmark & \xmark & \xmark & \pmark & \xmark & \xmark & \xmark & \pmark & \xmark \\
HUMOTO~\cite{Lu_2025_HUMOTO} & 2025 & 2.2 & 1 & Opt & 0 & \xmark & \xmark & \xmark & \xmark & \cmark & \cmark & \xmark & \xmark & \pmark & \pmark \\
\midrule
\rowcolor{datasetcategory}
\multicolumn{16}{l}{\textbf{\textit{Egocentric body motion}}} \\
\midrule
EgoCap~\cite{rhodin2016egocap} & 2016 & 0.7 & 8 & MV & 2 & \cmark & \xmark & \xmark & \xmark & \xmark & \xmark & \xmark & \xmark & \xmark & \xmark \\
Mo$^2$Cap$^2$~\cite{xu2019mo} & 2019 & 530K fr & 2 & Syn/MV & 1 & \cmark & \xmark & \xmark & \xmark & \xmark & \xmark & \xmark & \xmark & \xmark & \xmark \\
xR-EgoPose~\cite{tome2019xr} & 2019 & 380K fr & \na & Syn & 1 & \cmark & \xmark & \xmark & \xmark & \pmark & \xmark & \xmark & \xmark & \xmark & \xmark \\
HPS~\cite{HPS} & 2021 & 3+ & 7 & IMU & 1 & \xmark & \xmark & \xmark & \cmark & \xmark & \xmark & \cmark & \xmark & \pmark & \xmark \\
EgoBody~\cite{zhang2022egobody} & 2022 & 220K fr & 36 & MV & 1 & \xmark & \xmark & 3--5 & \xmark & \pmark & \xmark & \cmark & \xmark & \xmark & \xmark \\
ADT~\cite{pan2023aria} & 2023 & 6.6 & \na & MV & 3 & \xmark & \xmark & \xmark & \cmark & \xmark & \cmark & \cmark & \cmark & \xmark & \pmark \\
UnrealEgo2~\cite{akada20243d} & 2024 & 2.5M fr & 17 & Syn & 2 & \cmark & \xmark & \xmark & \xmark & \pmark & \xmark & \pmark & \xmark & \pmark & \xmark \\
UnrealEgo-RW~\cite{akada20243d} & 2024 & 1.4 & 16 & MV & 2 & \cmark & \xmark & \xmark & \xmark & \xmark & \xmark & \pmark & \xmark & \pmark & \xmark \\
EgoWholeBody~\cite{wang2024egocentric} & 2024 & 870K fr & 14 & Syn & 1 & \cmark & \xmark & \xmark & \xmark & \cmark & \xmark & \xmark & \xmark & \pmark & \xmark \\
Ego-Exo4D~\cite{grauman2024ego} & 2024 & 1286 & 740 & Ps & 3 & \xmark & \xmark & 4--5 & \cmark & \pmark & \xmark & \cmark & \cmark & \pmark & \pmark \\
Nymeria~\cite{ma2024nymeria} & 2024 & 300 & 264 & IMU & 3 & \xmark & 2 & 1 & \cmark & \xmark & \xmark & \pmark & \cmark & \pmark & \pmark \\
EgoBody3M~\cite{zhao2024egobody3m} & 2024 & 30+ & 120 & MV & 4 & \xmark & \xmark & \xmark & \xmark & \xmark & \xmark & \xmark & \xmark & \pmark & \xmark \\
EMHI~\cite{fan2025emhi} & 2025 & 28.5 & 58 & MV+IMU & 2 & \cmark & \xmark & \xmark & \cmark & \xmark & \xmark & \xmark & \xmark & \pmark & \xmark \\
Ego-Elec~\cite{zhu2025human} & 2025 & 9 & \na & IMU & 3 & \cmark & \xmark & \xmark & \cmark & \xmark & \xmark & \xmark & \xmark & \pmark & \xmark \\
FRAME~\cite{camiletto2025frame} & 2025 & 7.4 & 16 & MV & 2 & \cmark & \xmark & \xmark & \xmark & \xmark & \xmark & \xmark & \xmark & \pmark & \xmark \\
Eva-3M~\cite{dai2026egocentric} & 2026 & 3.0M fr & 31 & MV & 2 & \cmark & \xmark & \xmark & \xmark & \xmark & \xmark & \xmark & \xmark & \pmark & \xmark \\
OmniEgoDB~\cite{cho2026omniegocap} & 2026 & 3 & 4 & IMU & 1--3 & \xmark & \xmark & \na & \cmark & \pmark & \xmark & \xmark & \na & \pmark & \na \\
\midrule
\rowcolor{datasetcategory}
\multicolumn{16}{l}{\textbf{\textit{Human data for robot learning}}} \\
\midrule
Humanoid-X~\cite{mao2025universal} & 2025 & 240.3 & \na & Ps & 0 & \xmark & \xmark & \na & \xmark & \xmark & \xmark & \xmark & \xmark & \pmark & \xmark \\
PHUMA~\cite{lee2026phuma} & 2026 & 73.0 & \na & Opt+Ps & 0 & \xmark & \xmark & \xmark & \xmark & \xmark & \xmark & \xmark & \xmark & \cmark & \xmark \\
EgoDex~\cite{hoque2026egodex} & 2026 & 829 & \na & \xmark & 1 & \xmark & \xmark & \xmark & \na & \pmark & \xmark & \xmark & \xmark & \xmark & \xmark \\
OmniRetarget~\cite{yang2026omniretarget} & 2026 & $\sim$9 & \na & Opt & 0 & \xmark & \xmark & \xmark & \xmark & \xmark & \pmark & \pmark & \xmark & \pmark & \cmark \\
BONES-SEED~\cite{luo2026sonic} & 2026 & 288 & 522 & Opt & 0 & \xmark & \xmark & \xmark & \xmark & \na & \xmark & \xmark & \xmark & \cmark & \pmark \\
EgoHTR~\cite{brandes2026egohtr} & 2026 & 1.37 & 8 & IMU & 3 & \xmark & \xmark & 2 & \cmark & \xmark & \xmark & \cmark & \cmark & \cmark & \xmark \\
EgoVerse~\cite{punamiya2026egoverse} & 2026 & 1362 & 2087 & \xmark & 1+ & \xmark & \xmark & \xmark & \na & \pmark & \xmark & \xmark & \na & \xmark & \xmark \\
HumanVerse-500~\cite{xu2026towards} & 2026 & 500 & \na & Ps & 1 & \xmark & \xmark & \xmark & \na & \pmark & \xmark & \na & \na & \pmark & \cmark \\
Gen-HumanWholeBody~\cite{genrobot2026humanwholebody} & 2026 & 600 & \na & Ps & 6 & \na & \na & \na & \na & \pmark & \na & \na & \na & \pmark & \cmark \\
EgoExplorer~\cite{zhou2026egoexplorer} & 2026 & 100 & \na & Ps & 1 & \xmark & \xmark & \xmark & \na & \cmark & \na & \cmark & \na & \pmark & \cmark \\
ACE-Data-0~\cite{cao2026ace} & 2026 & 150 & 50 & Opt & 4 & \cmark & \xmark & 8 & \cmark & \cmark & \cmark & \xmark & \cmark & \pmark & \cmark \\
\midrule
\rowcolor{datasetours}
\textbf{RoboMoCap (Ours)} & \textbf{2026} & \textbf{200+} & \textbf{14} & \textbf{Opt} & \textbf{6} & \cmark & \textbf{2} & \textbf{1} & \cmark & \pmark & \cmark & \pmark & \cmark & \cmark & \cmark \\
\bottomrule
\end{tabular}}
\end{table*}
}

\newcommand{\hardwaresetup}{
\begin{figure*}
    \centering   \includegraphics[width=\textwidth]{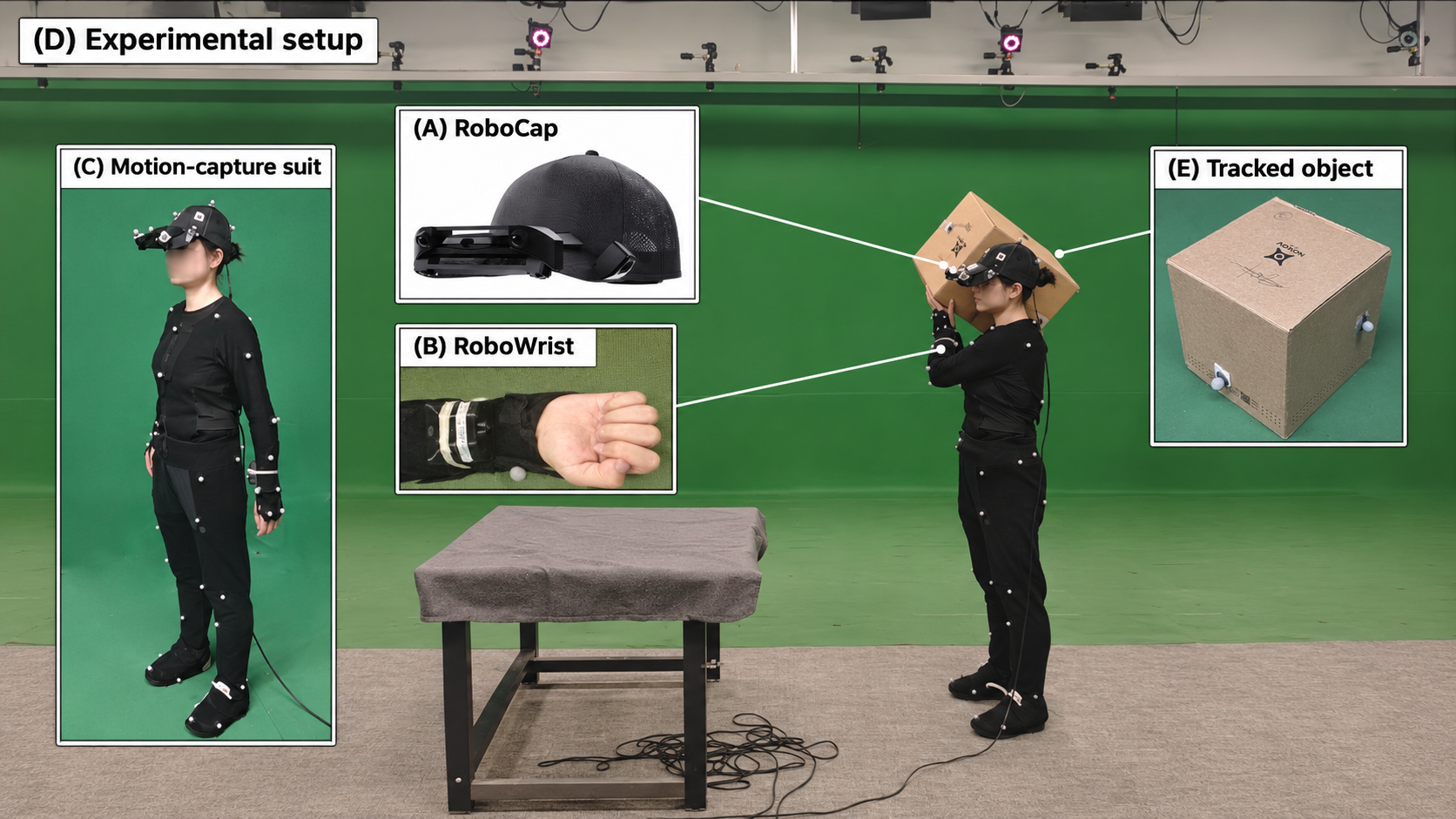}
    \vspace{-15pt}
    \caption{\textbf{RoboMoCap hardware setup.} (A) RoboCap~\cite{ali2026robocapnewplatformegocentric} for head-mounted egocentric sensing. (B) RoboWrist~\cite{bitrobot2026robocap} for wrist-mounted visual capture. (C) Motion capture suit with reflective markers for NOKOV optical tracking. (D) Recording studio equipped with NOKOV optical tracking cameras. (E) Example rigid object fitted with reflective markers for 6-DoF pose tracking.}
    \vspace{-5pt}
    \label{fig:hardwaresetup}
\end{figure*}
}

\newcommand{\dataprocessingpipeline}{
\begin{figure*}
    \centering   \includegraphics[trim={0cm 0cm 0cm 0cm},clip,width=\textwidth]{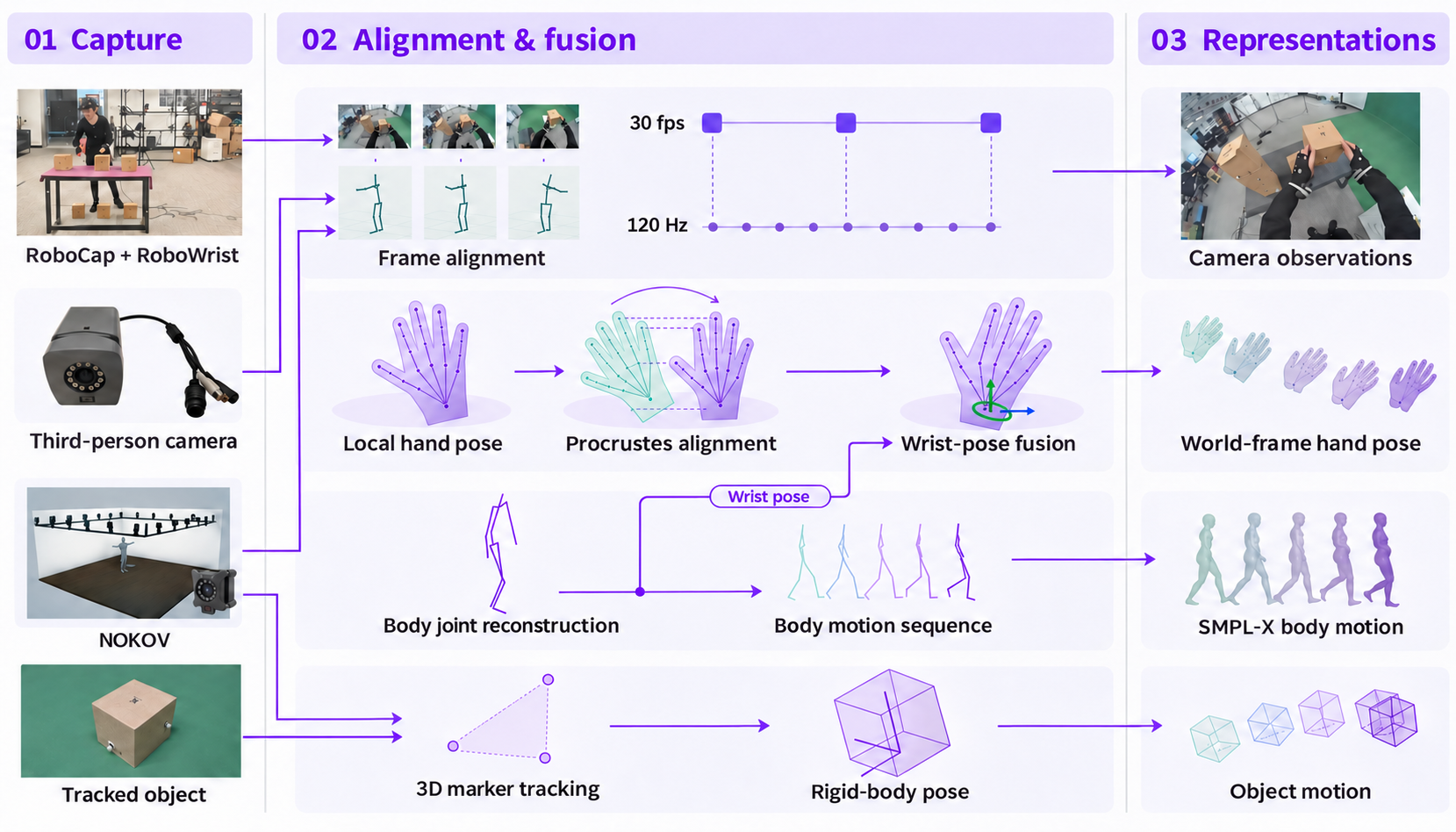}
    \vspace{-15pt}
    \caption{\textbf{Overview of the RoboMoCap capture and annotation pipeline.}
    \textbf{(1) Capture:} A multimodal recording system combines
    head-mounted and wrist-mounted cameras, third-person video,
    and NOKOV optical motion capture to record synchronised
    egocentric observations, full-body motion, and object
    interactions.
    \textbf{(2) Alignment and fusion:} Visual observations and
    optical motion capture are temporally aligned across
    30\,fps video and 120\,Hz motion measurements. Local
    hand-pose estimates are aligned and fused with tracked
    wrist poses, while body joint trajectories and rigid-object
    poses are reconstructed from optical MoCap.
    \textbf{(3) Representations:} The resulting annotations
    provide calibrated camera observations, world-frame
    hand poses, SMPL-X body motion sequences, and tracked
    object motion for egocentric whole-body motion
    understanding and humanoid learning.}
    \vspace{-5pt}
    \label{fig:dataprocessingpipeline}
\end{figure*}
}

\newcommand{\datasetanalysis}{
\begin{figure*}
    \centering   \includegraphics[trim={0cm 0cm 0cm 0cm},clip,width=\textwidth]{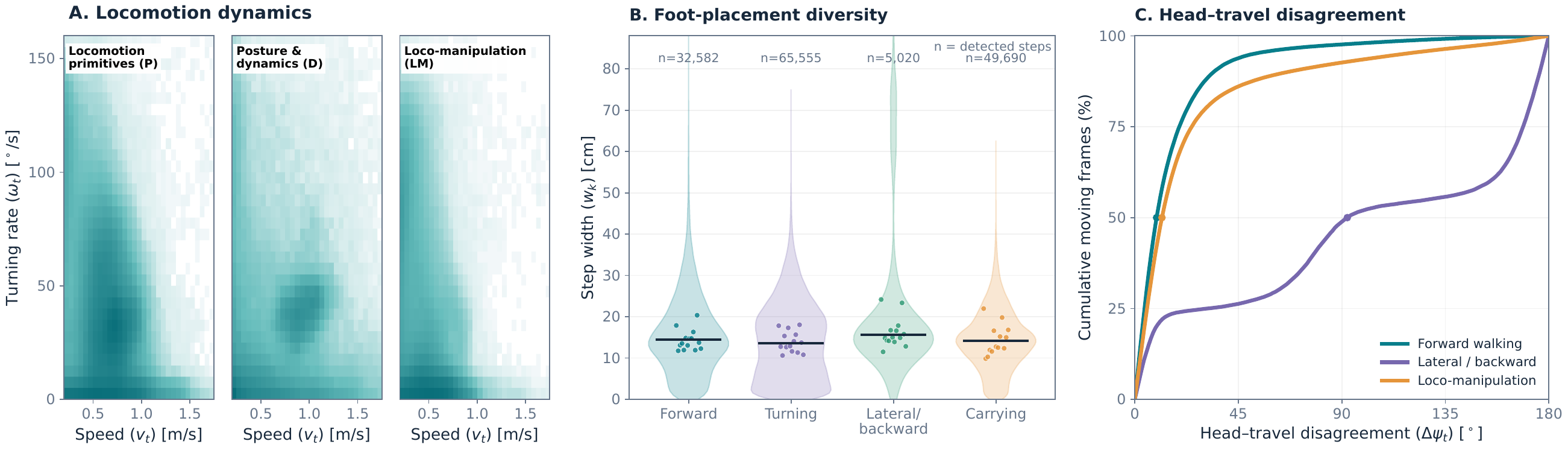}
    \vspace{-15pt}
    \caption{\textbf{Locomotion and loco-manipulation diversity in RoboMoCap.}
    (A) Distributions of pelvis translational speed and turning
    rate across locomotion primitives, posture and dynamics,
    and loco-manipulation activities.
    (B) Step-width distributions across forward walking,
    turning, lateral/backward walking, and object carrying.
    Violin widths indicate step density, horizontal lines
    denote medians, and dots represent participant-level
    medians ($n$ denotes detected steps).
    (C) Cumulative distributions of angular disagreement
    between head orientation and travel direction across
    movement conditions, highlighting the challenges of
    inferring locomotion from egocentric observations.}
    \vspace{-5pt}
    \label{fig:datasetanalysis}
\end{figure*}
}

\newcommand{\actiondistribution}{
\begin{figure}[t]
    \centering
    \includegraphics[trim={0cm 0cm 12cm 0cm},clip, width=\columnwidth]{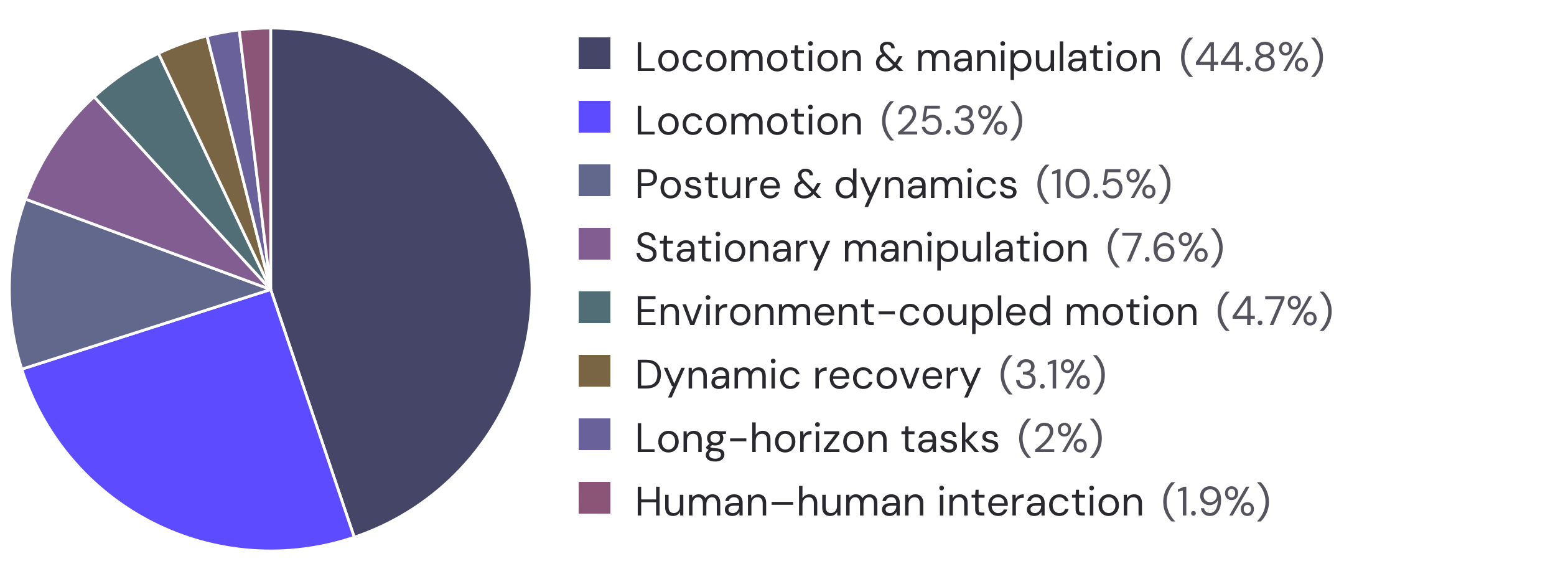}
    \vspace{-15pt}
    \caption{\textbf{Distribution of recorded data across action categories.} RoboMoCap emphasises locomotion and loco-manipulation while covering diverse whole-body activities.}
    \vspace{-15pt}
    \label{fig:actiondistribution}
\end{figure}
}


\twocolumn[{%
    \renewcommand\twocolumn[1][]{#1}%
    \maketitle

    \begin{center}
        \includegraphics[trim={0cm 0cm 0cm 0cm},
            clip,
            width=\textwidth,
            height=\textwidth,
            keepaspectratio
        ]{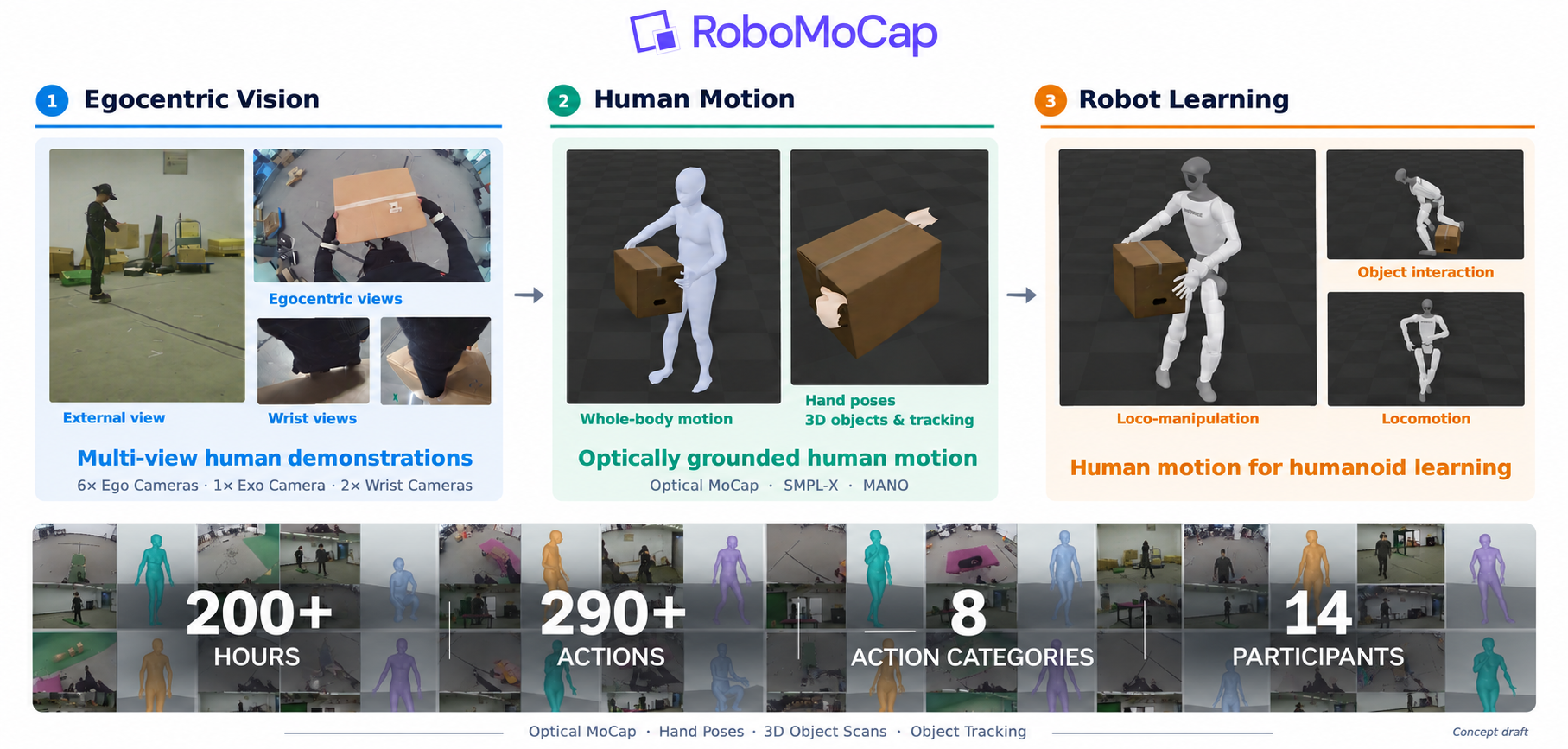}
        \captionof{figure}{\textbf{RoboMoCap: From Human Demonstrations to Robot Learning.}
        RoboMoCap connects egocentric vision with
        whole-body motion for humanoid robot learning.
        (1) \textbf{Egocentric Vision} captures synchronised
        multi-view observations from head-mounted,
        wrist-mounted, and exocentric cameras.
        (2) \textbf{Human Motion} provides optically captured
        whole-body motion, articulated hand poses, and
        tracked 3D objects.
        (3) \textbf{Robot Learning} leverages diverse
        locomotion, manipulation, and loco-manipulation
        demonstrations to support humanoid robot learning.
        The dataset comprises over 200 hours of
        recordings from 14 participants, covering over
        290+ actions across eight categories.}
        \label{fig:teaser}
    \end{center}
    \vspace{0.5em}}]

\begin{abstract}
Humanoid robots must coordinate locomotion and manipulation to navigate complex environments and perform everyday tasks. Learning such behaviours from human demonstrations requires accurate whole-body motion paired with egocentric vision, particularly for lower-body movements and foot placements that are frequently invisible from a first-person perspective. However, existing egocentric datasets often lack high-fidelity optical motion capture or systematic coverage of diverse loco-manipulation behaviours. To address this gap, we introduce RoboMoCap, a large-scale multimodal dataset combining egocentric vision with marker-based optical motion capture, specifically designed for locomotion-centric humanoid learning. RoboMoCap comprises over 200 hours of recordings from 14 participants, featuring six head-mounted cameras, wrist-mounted visual and inertial sensing, and optical full-body motion capture at 120 Hz. The dataset systematically covers diverse locomotion primitives, dynamic whole-body movements, and loco-manipulation tasks, including variable-speed walking, turning, constrained navigation, terrain traversal, object transportation, and manipulation during locomotion. It provides temporally aligned multimodal observations and metric whole-body motion representations, enabling the study of human movement beyond directly observable body regions. We systematically evaluate existing methods on three benchmark tasks: egocentric full-body motion reconstruction, foot contact and placement estimation, and human-to-humanoid motion retargeting. 
By pairing egocentric vision with high-fidelity human motion across diverse locomotion and interaction scenarios, RoboMoCap provides a foundation for egocentric motion understanding and humanoid learning from human demonstrations. 
\end{abstract}

\section{Introduction}
\label{sec:intro}

Humanoid robots are increasingly expected to operate in human environments, where successful task execution requires the seamless coordination of locomotion, balance, and manipulation ~\cite{h2o,beyondmimic,sonic,jia2026reflex}. Everyday activities, such as carrying objects across a room, navigating narrow passages, stepping over obstacles, or approaching a table to place an object, demand continuous coordination between the upper and lower body. Rather than treating locomotion and manipulation as independent skills, these behaviours require adapting foot placement, body posture, and movement trajectories to the surrounding environment and ongoing interactions. Learning such coordinated behaviours from human demonstrations offers a promising direction for developing versatile humanoid robots. However, this approach fundamentally depends on datasets that capture not only what humans observe and interact with, but also how their entire bodies move throughout these activities.

Egocentric sensing provides a natural perspective for learning from human behaviour, capturing visual observations closely aligned with the information available to an embodied agent. Large-scale egocentric datasets have demonstrated the value of first-person recordings for activity understanding, human-object interaction, and robot learning. Nevertheless, recovering precise whole-body motion from egocentric observations remains challenging. In particular, the lower body is frequently outside the field of view of head-mounted cameras, making foot placement, support transitions, and lower-body articulation difficult to infer directly. Furthermore, similar head trajectories may correspond to substantially different locomotion patterns, while head orientation does not necessarily align with body heading or travel direction. These ambiguities become particularly pronounced during turning, lateral or backward walking, dynamic posture transitions, and object transportation. Consequently, connecting egocentric perception with accurate whole-body motion supervision is essential for understanding locomotion and its interaction with manipulation.

Existing datasets capture complementary aspects of human motion. AMASS~\cite{AMASS:ICCV:2019} provides detailed human kinematics but lacks synchronised egocentric observations. Nymeria~\cite{ma2024nymeria} pairs wearable first-person sensing with full-body motion, while EgoHTR~\cite{brandes2026egohtr} focuses on challenging terrain traversal. More recently, ACE-Data-0~\cite{cao2026ace} combines egocentric observations with optical motion capture, and Ego-Exo4D-HM~\cite{maddukuri2026egoexo4dhumanmeshesdataset} provides dense motion reconstructions from multi-view recordings. However, datasets combining synchronised egocentric observations and high-fidelity optical whole-body supervision across diverse locomotion and loco-manipulation behaviours remain limited. Table~\ref{tab:dataset_comparison} situates RoboMoCap within this broader landscape, comparing sensing modalities, motion supervision, and behavioural coverage.

To bridge this gap, we introduce \textit{RoboMoCap}, a large-scale multimodal human motion dataset designed for locomotion-centric humanoid learning. RoboMoCap combines a custom six-camera head-mounted sensing platform with wearable wrist sensors and a marker-based optical motion capture system. The wearable cameras record complementary scene-facing and body-facing perspectives, while optical motion capture measures full-body motion at 120 Hz. Through temporal synchronisation and spatial calibration, we associate egocentric observations with metric body motion, tracked camera trajectories, and fitted SMPL-X~\cite{SMPL-X:2019} representations. Crucially, marker-based optical motion capture provides whole-body supervision independently of egocentric visual inference, enabling direct evaluation of lower-body motion even when the corresponding body regions are not visible in the wearable cameras.

A central design objective of RoboMoCap is to capture human locomotion beyond conventional forward walking. To this end, our collection protocol systematically organises activities into locomotion primitives, posture and whole-body dynamics, and loco-manipulation. Locomotion tasks span variations in speed, movement direction, turning, obstacle negotiation, and terrain conditions, while dynamic activities introduce changes in posture, support configuration, and movement intensity. Loco-manipulation tasks further couple locomotion with object acquisition, transportation, and placement under varying loads and environmental constraints. Together, these activities capture both isolated motion primitives and coordinated whole-body behaviours, enabling systematic analysis of how locomotion adapts to environmental conditions and concurrent manipulation demands.

Beyond providing synchronised multimodal recordings, RoboMoCap enables systematic evaluation of egocentric whole-body motion understanding and human-to-humanoid motion transfer. We evaluate existing methods across three benchmark tasks: egocentric full-body motion reconstruction, foot contact and placement estimation, and human-to-humanoid motion retargeting. These evaluations examine the reconstruction of whole-body kinematics, the accuracy and physical plausibility of lower-body motion, and the preservation of human motion characteristics when transferred to humanoid embodiments. Together, the dataset and benchmarks provide a resource for investigating egocentric motion understanding and its potential applications in humanoid learning.

Our main contributions are summarised as follows:
\begin{itemize}
\item \textbf{The largest locomotion-centric multimodal human motion dataset.} We introduce RoboMoCap, combining synchronised egocentric visual and inertial sensing with marker-based optical full-body motion capture at 120 Hz, with a focus on diverse locomotion and loco-manipulation behaviours.

\item \textbf{A systematic locomotion and loco-manipulation collection protocol.} We design a structured action inventory spanning diverse gait patterns, movement directions, posture transitions, terrain conditions, and object interactions during locomotion, providing calibrated camera trajectories and metric whole-body motion representations.

\item \textbf{Benchmarks for egocentric motion understanding and humanoid retargeting.} We evaluate existing methods on egocentric full-body motion reconstruction, foot contact and placement estimation, and human-to-humanoid motion retargeting, establishing reference results for studying lower-body motion accuracy and cross-embodiment motion transfer.

\end{itemize}
\section{Related Work}
\label{sec:relatedworks}

\datasetcomparison

\subsection{Motion Capture Datasets}
Motion capture provides high-quality references for
human motion modelling and synthesis.
AMASS~\cite{AMASS:ICCV:2019} unifies marker-based
datasets under a common body-model parameterisation,
while LAFAN1~\cite{harvey2020robust} and
100STYLE~\cite{mason2022real} offer diverse locomotion
and stylised gait sequences. Interaction-oriented
datasets such as GRAB~\cite{taheri2020grab},
OMOMO~\cite{li2023object}, and
HUMOTO~\cite{Lu_2025_HUMOTO} extend motion capture
to whole-body object manipulation.
Although these datasets provide detailed body motion,
they generally lack corresponding real egocentric
observations, limiting their direct use for learning
vision-conditioned humanoid behaviours. Moreover,
locomotion and object manipulation are typically
studied separately rather than through systematic
variations of their coordination.

\subsection{Egocentric Human Motion Datasets}
Early egocentric pose datasets employ body-facing
cameras to observe the wearer.
EgoCap~\cite{rhodin2016egocap} and
Mo$^2$Cap$^2$~\cite{xu2019mo} explore this setting,
while xR-EgoPose~\cite{tome2019xr},
UnrealEgo2~\cite{akada20243d}, and
EgoWholeBody~\cite{wang2024egocentric} provide
synthetic training data. Real-world datasets such
as UnrealEgo-RW~\cite{akada20243d},
EMHI~\cite{fan2025emhi},
Ego-Elec~\cite{zhu2025human}, and
Eva-3M~\cite{dai2026egocentric} further support
egocentric body reconstruction.

Scene-facing datasets, including
EgoBody~\cite{zhang2022egobody},
HPS~\cite{HPS}, and ADT~\cite{pan2023aria},
capture human activities in their environmental
context. EgoBody3M~\cite{zhao2024egobody3m},
Ego-Exo4D~\cite{grauman2024ego},
Nymeria~\cite{ma2024nymeria}, and NymeriaPlus \cite{detone2026nymeriaplus} extend real-world
egocentric motion capture through larger-scale
recordings and alternative body supervision. FRAME~\cite{camiletto2025frame} further addresses
real-world egocentric body reconstruction, while
EgoSim ~\cite{hollidt2024egosim}
and EgoWholeView~\cite{hu2026egowholeview}
explore multi-camera wearable configurations
for whole-body motion capture. More recently, EgoHTR~\cite{brandes2026egohtr}
targets terrain traversal, while
ACE-Data-0~\cite{cao2026ace} combines egocentric
observations with optical motion capture for
long-horizon household activities. Despite these
advances, large-scale optical supervision with
systematic coverage of locomotion dynamics,
foot placement, and loco-manipulation remains
relatively limited.

\subsection{Human Motion for Humanoid Learning}
Human motion datasets increasingly support humanoid
control through retargeting and imitation.
Humanoid-X~\cite{mao2025universal} derives
240\,h of motion from internet videos, while
PHUMA~\cite{lee2026phuma} curates 73\,h of
physically refined locomotion and
BONES-SEED~\cite{luo2026sonic} provides 288\,h
of motion capture for humanoid training.
OmniRetarget~\cite{yang2026omniretarget}
further enables interaction-preserving retargeting
involving objects and terrain. However, these
motion-oriented datasets generally lack paired
egocentric observations.

Conversely, egocentric datasets for robot learning, such as EgoDex~\cite{hoque2026egodex} and EgoVerse~\cite{punamiya2026egoverse}, provide
large-scale human manipulation demonstrations,
while HumanVerse-500~\cite{xu2026towards} and
EgoExplorer~\cite{zhou2026egoexplorer}
extend first-person observations to whole-body
loco-manipulation using estimated human motion.
RoboMoCap bridges these directions by pairing
egocentric observations with optical whole-body
motion across diverse locomotion and
loco-manipulation tasks, supporting egocentric
motion understanding, humanoid retargeting,
and vision-conditioned humanoid learning.

\hardwaresetup

\section{RoboMoCap Capture System}
\label{sec:capturesystem}

\subsection{Hardware Setup}
\label{sec:hardwaresetup}

\noindent\textbf{Egocentric Cameras.}
All participants wear a RoboCap \cite{ali2026robocapnewplatformegocentric}, a lightweight head-mounted sensing platform equipped with six hardware-synchronised global-shutter RGB cameras, two inertial measurement units (IMUs), and a magnetometer (\cref{fig:hardwaresetup}-(A)). The camera configuration comprises
two fisheye stereo pairs (front-facing and downward-facing eye cameras) and two lateral monocular wide-angle cameras, providing complementary views of the surrounding environment, hands, and body. All six cameras record at a resolution of $1920\times1080$ pixels and $30$~Hz, with hardware-triggered synchronisation. The two IMUs record accelerometer and gyroscope measurements at $200$~Hz, while the auxiliary magnetometer operates at $100$~Hz. \\
\noindent\textbf{Wrist-mounted Cameras.} Participants additionally wear a RoboWrist \cite{bitrobot2026robocap} on each wrist (\cref{fig:hardwaresetup}-(B)). Each device integrates a wide-angle RGB camera, an IMU, and a magnetometer, providing complementary hand-centric observations during locomotion and object interaction. The cameras record at $1920 \times 1080$ pixels and $30$~Hz, while the IMUs capture accelerometer and gyroscope measurements at $200$~Hz and the magnetometers operate at $100$~Hz. Both RoboWrist devices are paired with RoboCap for coordinated recording, with their independently clocked streams aligned during subsequent data processing. \\
\noindent\textbf{Exocentric Camera.} A single exocentric RGB camera captures a third-person view of participants within the motion capture studio, providing complementary visual observations of whole-body movements and interactions with the surrounding environment. The camera records at a resolution of $1920\times1080$ pixels and $30$~Hz. \\
\noindent\textbf{Optical Motion Capture System.} To capture high-fidelity whole-body motion independently of egocentric visibility, each participant wears a motion-capture suit equipped with 53 reflective markers (\cref{fig:hardwaresetup}-(C)), tracked using a NOKOV optical motion capture system \cite{nokov2026mocap}. The system employs multiple infrared cameras distributed around the recording area to capture three-dimensional marker trajectories at $120$~Hz (\cref{fig:hardwaresetup}-(D)). Additionally, four reflective markers are rigidly attached to the RoboCap headset, forming an independently tracked rigid body, denoted as \textit{Tracker0}. This configuration enables 6-DoF tracking of the headset in the optical motion capture coordinate system, providing a reference for temporal synchronisation and spatial registration between the wearable sensors and optical motion capture. \\
\noindent\textbf{Object Tracking.} We first acquire a 3D scan of each rigid object using the LiDAR-based Reality Composer app~\cite{apple2023realitycomposer} on an iPhone 14 Pro. We then attach four reflective markers to each object (\cref{fig:hardwaresetup}-(E)), enabling the NOKOV optical motion capture system to track its 6-DoF pose at $120$~Hz. This setup provides metric object trajectories throughout manipulation and transportation tasks, allowing analysis of object motion alongside synchronised whole-body movements. \\ 
\noindent\textbf{Audio.}
A microphone attached to the RoboCap records
audio during data collection, capturing audible cues associated with contact events, spoken instructions, and ambient environmental sounds. 

\subsection{Capture Environment}

A central design consideration of RoboMoCap is to capture diverse whole-body movements while maintaining reliable optical motion tracking. Data collection is conducted in NOKOV's motion capture studio, comprising a $7 \times 7$ m recording area (\cref{fig:hardwaresetup}-(D)). Unlike conventional manipulation-centred capture setups \cite{hoque2026egodex, punamiya2026egoverse, taheri2020grab} that primarily focus on stationary workspaces, our recording environment accommodates continuous locomotion, changes in movement direction, dynamic postural transitions, and interactions with surrounding objects. The recording area is configured to support a wide range of locomotion behaviours, including straight-line walking, turning, circular and figure-eight trajectories, and repeated shuttle movements. Task-specific arrangements introduce spatial constraints and environmental variations, including narrow passages, obstacles, inclined surfaces, and changes in elevation. For loco-manipulation activities, interaction areas are arranged to accommodate object pickup, transportation, and placement at different locations and heights. Objects with varying physical properties and carrying requirements are incorporated to capture the coordination between locomotion, posture, and manipulation. These configurations allow both controlled motion primitives and continuous whole-body activities to be recorded within a common sensing environment.

\section{RoboMoCap Dataset}

\subsection{Design Principles and Action Taxonomy}
\label{sec:design_principles}

\taxonomy
\dataprocessingpipeline

RoboMoCap is designed to support egocentric whole-body motion understanding and humanoid robot learning, with a particular emphasis on locomotion and its coordination with manipulation. We adopt four key design principles to guide data collection.

\noindent \emph{\textbf{(i)~Locomotion-centric diversity:}}  Human locomotion involves substantially more than steady forward walking. Everyday movement requires changes in speed, direction, body orientation, and foot placement to accommodate environmental constraints. We therefore emphasise diverse locomotion behaviours,
including turning, lateral and backward walking, obstacle avoidance, terrain traversal, and dynamic postural transitions. This systematic coverage aims to capture a broad distribution of lower-body movements relevant to humanoid locomotion. \\
\emph{\textbf{(ii)~Egocentric motion observability:}}
A fundamental challenge in egocentric motion reconstruction is that the lower body, particularly the feet, is frequently outside the camera's field of view. Moreover, similar head movements can correspond to different body configurations and foot placements. We design activities that expose these ambiguities, including changes in gaze, walking direction, and body orientation. By pairing egocentric observations with optical whole-body motion measurements, RoboMoCap provides supervision for both visible and unobserved body movements. \\
\emph{\textbf{(iii)~Locomotion--manipulation coordination:}}
Humanoid robots must coordinate locomotion with object interaction rather than execute these behaviours independently. We therefore incorporate tasks involving object pickup, transportation,
placement, and manipulation during locomotion. Variations in carrying configuration, object load, and navigation constraints capture how upper-body activities influence whole-body coordination
and locomotion strategies. \\
\emph{\textbf{(iv)~Structured and continuous behaviours:}}
We combine controlled motion primitives with extended, multi-stage activities to capture both individual movement patterns and their
temporal transitions. Structured task variations enable comparisons across movement conditions, while continuous recordings preserve natural transitions between locomotion, postural adjustments, and object interactions. This design supports both fine-grained motion analysis and longer-horizon embodied tasks. \\
\noindent \textbf{Action Taxonomy.} 
To systematically capture diverse whole-body behaviours, we organise RoboMoCap around three core action categories: locomotion primitives, posture and whole-body dynamics, and loco-manipulation. Beyond these core protocols, the recorded dataset also includes static manipulation, environment-coupled whole-body activities, dynamic recovery, human--human loco-manipulation, and long-horizon composite tasks. These additional activities extend the dataset towards more complex whole-body interactions and multi-stage behaviours.  
~\cref{tab:action_taxonomy} summarises the taxonomy and its principal variations. The collection protocol comprises 291 predefined tasks, systematically varying locomotion patterns and body configurations.

\subsection{Collection Protocol}
\label{sec:collection_protocol}

Data collection follows predefined protocols specifying task instructions, scene configurations, repetitions, and movement variations. Participants are instructed to move naturally at a comfortable pace without imitating robotic movements. Task conditions are systematically varied through movement direction, path geometry, obstacle arrangements, object loads, and interaction targets. Left- and right-sided variations are explicitly instructed to encourage balanced coverage.

Repeated movements are recorded in continuous takes rather than isolated repetitions, preserving natural transitions between motion phases. Loco-manipulation tasks capture object acquisition, transport, and placement, including activities requiring simultaneous walking and manipulation. To isolate the influence of locomotion, selected static manipulation tasks are paired with corresponding loco-manipulation protocols, using the same participant and objects. Long-horizon tasks specify initial scene configurations and goals while allowing participants to determine their own execution strategies.

Each take is associated with participant and action identifiers, task conditions, and execution metadata. Objects are reset between takes, while carried loads and reaching heights are adjusted to individual
comfort. Equipment checks and task-specific safety procedures are incorporated for dynamic movements and recovery activities.

\subsection{Synchronisation and Calibration}

\noindent \textbf{Temporal Synchronisation.} Synchronising RoboCap's visual-inertial recordings with the optical motion capture system is essential for establishing accurate correspondences between egocentric observations and whole-body motion. Since the two systems operate on independent clocks, we estimate their temporal alignment by exploiting the head motion observed simultaneously by RoboCap's IMU and the optically tracked rigid body (Tracker0) attached to the headset.

We first construct a reference timebase for the optical motion capture system using its frame indices and the recording frequency of $120$~Hz, with the origin defined at the first successfully solved body frame. We then compute angular-speed magnitudes from the RoboCap gyroscope and the temporal derivatives of Tracker0's orientation trajectory. By comparing angular-speed magnitudes rather than rotation vectors, the alignment is invariant to the relative orientation between the IMU and optical tracker, eliminating the need for their spatial calibration during temporal alignment. We clip both signals at $800^\circ/\mathrm{s}$ to suppress spurious peaks and estimate the initial clock offset by maximising their temporal cross-correlation.

To improve robustness against periodic movements and ambiguous correlation peaks, we repeat the alignment over $20$-second temporal windows and aggregate the estimated offsets using their median. We additionally examine the evolution of window-wise offsets to detect clock drift. When a statistically significant trend exceeds $3$~ms over the recording, we estimate an affine time mapping; otherwise, we retain an offset-only model. For each session $s$, the resulting transformation is

\begin{equation}
    t_{\mathrm{mocap}}
    = a_s t_{\mathrm{rig}} + b_s,
    \label{eq:temporal_sync}
\end{equation}

\noindent
where $t_{\mathrm{rig}}$ denotes session-relative RoboCap time, $b_s$ is the estimated temporal offset, and $a_s$ accounts for relative clock-rate differences ($a_s=1$ in the absence of measurable drift).

To obtain frame-level correspondence, we recover the timestamp of each video frame from the recording's device-clock start time, its presentation timestamp (PTS), and the camera--IMU temporal offset obtained during device calibration. Discontinuities in video timestamps are examined using consistency between image-derived motion and gyroscope measurements to distinguish dropped frames from timestamp artefacts. Finally, Eq.~\ref{eq:temporal_sync} maps each frame
to the optical motion capture timebase. All modalities retain their native sampling rates and individual timestamps, without temporal resampling or video
re-encoding. This provides a common timeline for the aligned egocentric observations, inertial measurements, and optical body motion.

\noindent\textbf{Spatial Calibration.}
To establish spatial correspondence between the wearable cameras and optical motion capture, we use the per-device camera calibration provided by RoboCap
\cite{ali2026robocapnewplatformegocentric}. This includes camera intrinsics and inter-camera extrinsics, with camera projection modelled using the Kannala--Brandt \cite{kannala2004generic} fisheye distortion model.

To register the egocentric cameras with the optical motion capture coordinate system, we establish a fixed rigid transformation between each camera and the optically tracked headset rigid body, \textit{Tracker0}. The world-frame pose of camera $C_k$ at time $t$ is obtained as

\begin{equation}
    T_{WC_k}(t) = T_{WS}(t) T_{SC_k},
    \label{eq:spatial_calibration}
\end{equation}

where $W$ denotes the canonical world coordinate system, $S$ represents the Tracker0 coordinate frame, $T_{WS}(t)$ is its optically tracked 6-DoF pose, and $T_{SC_k}$ is the calibrated rigid transformation from camera $C_k$ to Tracker0.

This registration provides metric camera trajectories
aligned with the optical whole-body motion, enabling
direct correspondence between egocentric observations
and 3D body annotations. The wrist-mounted and
exocentric cameras are calibrated separately, as
they are not rigidly attached to the head-mounted
tracking assembly.





\noindent \textbf{Coordinate Canonicalisation.} Raw optical observations are recorded in millimetres in the motion capture system's Y-up coordinate convention. We transform these measurements into a right-handed, Z-up world coordinate system expressed in metres, using a consistent coordinate conversion and block-level levelling transformation. The same canonical frame is used for optical marker trajectories, body representations, and registered camera poses. Importantly, the levelled world origin is not assumed to coincide exactly with the physical ground-contact surface; foot-contact analysis must account for marker attachment and support-surface geometry.

\subsection{Multimodal Annotation Generation}

Figure~\ref{fig:dataprocessingpipeline} illustrates the
RoboMoCap capture and annotation pipeline, which
integrates synchronised egocentric observations with
optical motion capture to generate consistent
representations of full-body motion, articulated
hand poses, and object trajectories.

\noindent\textbf{Whole-Body Motion Reconstruction.} To obtain a consistent parametric representation of human motion, we fit the SMPL-X body model~\cite{SMPL-X:2019} to the optical motion capture measurements. The fitting process leverages the articulated skeleton reconstructed by NOKOV for pose initialisation, while the observed 3D marker trajectories provide geometric constraints on the body surface.

We first estimate body shape from individual recording sessions and aggregate these estimates to obtain a consistent participant-specific shape. With the body shape fixed, we optimise the time-varying body pose and global translation, together with session-specific marker attachment offsets, to minimise discrepancies between observed marker positions and their corresponding locations on the fitted SMPL-X body.

The resulting SMPL-X parameters provide a consistent representation of articulated body motion, including global position, body orientation, joint articulation, and mesh geometry. Unlike motion estimated solely from egocentric observations, the fitting process is constrained by optical MoCap measurements of the entire body, including the legs and feet, which are frequently outside the wearable cameras' field of view. This enables the derivation of lower-body kinematics and foot trajectories for locomotion-oriented analysis and evaluation.

\datasetanalysis

\noindent \textbf{Hand Motion Estimation.} To complement the optical whole-body measurements with articulated hand motion, we use Grounded API
\cite{ali2026robocapnewplatformegocentric}. Its multi-view hand tracking model processes calibrated egocentric camera observations to estimate metric 3D hand poses represented using the MANO hand model \cite{MANO:SIGGRAPHASIA:2017}. By combining information from multiple camera views, the method estimates hand articulation under challenging egocentric conditions, including partial visibility and hand--object occlusion. Unlike the optically measured body markers, these hand annotations are vision-derived estimates. We therefore distinguish estimated finger articulation from optical body motion measurements in the released annotations.

To produce hand reconstructions that are consistent with the body meshes, we align Grounded API's local hand pose with the SMPL-X wrist pose. We fine-tune the model on a small training set (34.9 minutes), use the validation set to select the best checkpoint, and report the Procrustes-Aligned Mean Per-Joint Position Error (PA-MPJPE) on a test set (9.2 minutes). We use the same subjects for these splits as in the rest of the dataset.

\noindent \textbf{Scene Reconstruction.} 
We reconstruct the static 3D environment scene using MapAnything~\cite{keetha2026mapanything}, a feed-forward metric 3D reconstruction model. For this, we use the front-facing ego stereo cameras, together with their calibrated intrinsics and camera poses obtained from NOKOV optical tracking, to estimate scene depth. The predicted depth maps are back-projected into 3D and transformed into the NOKOV world coordinate system using the corresponding optically tracked camera poses. Consequently, the reconstructed scene geometry is directly expressed in the same coordinate frame as the captured human motion, without requiring any additional point-cloud alignment. 

\noindent \textbf{Task Annotations.} We provide detailed textual descriptions for each task, specifying the intended activity and its execution conditions. Each recording is additionally annotated with the start and end timestamps of individual task episodes, enabling the extraction of temporally localised action sequences.

\noindent \textbf{Data Quality.}
Each session passes automated checks on stream integrity, calibration and marker
labelling before release. Ego and mocap streams are aligned by cross-correlating the
head-IMU gyroscope with the angular speed of the head-mounted rigid body (median peak
correlation 0.985); offsets estimated in independent windows of a take deviate from the
per-take estimate by a median of 0.5\,ms (95th percentile 26\,ms). The fitted soles rest a median of 10\,mm above the floor (IQR 4--16\,mm), consistent with footwear on a barefoot body
model. Per-frame validity masks are provided for the body, marker, skeleton, head-rig, camera-pose and object streams.

\subsection{Dataset Statistics and Analysis}
\egocomparison

\kinematicfidelity
\kinematicfeasibility
\actiondistribution

\subsubsection{\textbf{Dataset Statistics.}}

\noindent \textbf{Action Distribution.} \cref{fig:actiondistribution} summarises the distribution of 200+ h across eight action categories. Locomotion primitives, posture and dynamics, and loco-manipulation collectively account for 161 hrs ($\approx80.6\%$) of the recordings, reflecting the dataset's emphasis on locomotion-centric whole-body behaviours.

\noindent\textbf{Participant and Recording Coverage.}
Participants contribute a median of 13.67\,h
(range: 9.79--21.95\,h), with the largest
contribution accounting for 11.4\% of the dataset.
Recording sessions have a median duration of
127\,s (IQR: 106.8--135.0\,s), preserving
continuous motion rather than isolated frames.

\noindent\textbf{Optical Annotation Quality.}
Optical capture achieves 99.81\% valid skeleton
frames and 99.74\% marker visibility. SMPL-X
fitting yields a median marker reconstruction
error of 4.78\,mm, demonstrating close agreement
between fitted body models and observed markers.

\subsubsection{Locomotion and Loco-Manipulation Diversity}
\label{sec:locomotion_diversity}

Recording duration and the number of action categories alone
do not fully characterise the diversity of human locomotion.
In particular, locomotion involves variations in translation,
rotation, foot placement, and the relationship between head
orientation and body movement. To quantify these properties,
we analyse the optical motion capture annotations across
locomotion primitives (P), posture and dynamics (D), and
loco-manipulation (LM) activities.
Figure~\ref{fig:datasetanalysis} summarises three complementary
aspects of the recorded motion. \\
\noindent\textbf{Locomotion Dynamics.}
Figure~\ref{fig:datasetanalysis}-(A) characterises the joint
distribution of pelvis translational speed and turning rate.
Given the pelvis position $\mathbf{p}_t$ and heading $\psi_t$,
we define horizontal speed and absolute turning rate as
\begin{equation}
v_t =
\frac{\|\mathbf{p}_{t+\Delta t}^{xy}
-\mathbf{p}_t^{xy}\|_2}{\Delta t},
\end{equation}
\begin{equation}
\omega_t =
\frac{|\operatorname{wrap}(
\psi_{t+\Delta t}-\psi_t)|}{\Delta t}.
\end{equation}
The analysis uses moving frames sampled at 10 Hz, with
speed and turning-rate bins of $0.05$ m/s and
$5^\circ$/s, respectively. Each heatmap displays
log-transformed bin counts normalised within its action
family, allowing the shape of the motion distribution
to be compared across categories.

The three action families exhibit different locomotion
profiles. Locomotion primitives have a median speed of
$0.65$ m/s and median turning rate of $27.1^\circ$/s,
whereas loco-manipulation activities have a lower median
speed of $0.45$ m/s and turning rate of $7.7^\circ$/s.
Nevertheless, loco-manipulation includes substantial
rotational motion, with a 95th-percentile turning rate
of $96.3^\circ$/s. These observations show that the
dataset captures diverse combinations of translation
and rotation, including slower object-related movement
and dynamic changes in heading. \\
\noindent\textbf{Foot-Placement Diversity.}
Figure~\ref{fig:datasetanalysis}-(B) examines step-width
distributions across four movement conditions:
forward walking, turning, lateral/backward walking,
and object carrying. Steps are extracted from the
optically tracked lower-body motion, and step width
is measured as the lateral separation of successive
foot placements in the pelvis-relative coordinate
frame. The violin plots represent the distributions
of individual detected steps, horizontal bars indicate
their medians, and dots show participant-level
median step widths.

We analyse 32,582 forward-walking steps, 65,555 turning
steps, 5,020 lateral/backward steps, and 49,690 carrying
steps. The corresponding median step widths are
$14.4$, $13.6$, $15.7$, and $14.1$ cm, respectively.
Lateral/backward locomotion exhibits the largest median
step width among these conditions, while all four groups
show substantial within-condition variation.

These distributions demonstrate that the dataset contains
diverse lower-body stepping patterns rather than only
repeated forward walking. Importantly, the recorded
foot placements provide motion-capture supervision
for movements where the feet may be outside the
egocentric cameras' field of view. This is particularly
relevant for evaluating foot-placement estimation,
support transitions, and whole-body reconstruction. \\
\noindent \textbf{Head--Travel Disagreement.}
Figure~\ref{fig:datasetanalysis}-(C) quantifies the
disagreement between the horizontal heading of the
headset and the actual direction of travel.
Let $\psi_t^{h}$ denote the headset heading and
$\psi_t^{v}$ the direction of horizontal motion.
Their absolute angular difference is
\begin{equation}
\Delta\psi_t =
\left|
\operatorname{wrap}
\left(\psi_t^{h}-\psi_t^{v}\right)
\right|.
\end{equation}
We compute the empirical cumulative distribution
of $\Delta\psi_t$ over moving frames, separately
for forward walking, lateral/backward walking,
and loco-manipulation activities.

The median disagreement is $9.5^\circ$ during forward
walking, indicating that head orientation and travel
direction are usually aligned. In contrast, the median
reaches $92.4^\circ$ during lateral/backward locomotion,
demonstrating that the wearer can travel in a direction
substantially different from the direction the headset
faces. Loco-manipulation exhibits a median disagreement
of $11.8^\circ$, but its 90th percentile reaches
$66.0^\circ$, indicating substantial disagreement in
a subset of moving frames.

These findings highlight a fundamental difficulty in
egocentric full-body reconstruction: head orientation
cannot be treated as a direct proxy for locomotion
direction. Even when head motion is accurately measured,
different lower-body movements and foot placements may
remain compatible with similar head observations.
By pairing egocentric recordings with optical full-body
motion capture, RoboMoCap provides supervision for
studying this ambiguity and evaluating motion
reconstruction under challenging locomotion conditions.

\section{Benchmarks}
\label{sec:experiments}

\subsection{Egocentric Full-Body Motion Reconstruction}
\label{sec:ego_reconstruction}

\lowerbodycomparison

\noindent\textbf{Task.}
We evaluate the ability of existing egocentric motion reconstruction methods to recover full-body human motion from ego observations. Unlike conventional pose estimation benchmarks, RoboMoCap provides optically grounded whole-body motion across diverse locomotion and loco-manipulation activities, enabling evaluation of both local articulation and global motion. We use MoCap-derived SMPL-X motion as the reference and evaluate publicly available pretrained methods without training or fine-tuning on RoboMoCap. \\
\noindent\textbf{Metrics.}
We evaluate joint position accuracy using Mean Per Joint Position Error in local (MPJPE) and world-frame (W-MPJPE), sequence-aligned MPJPE (WA-MPJPE) \cite{shen2024world}, and Procrustes-aligned MPJPE (PA-MPJPE) \cite{ionescu2013human3}. Joint rotation accuracy is measured using mean per-joint rotation error (MPJRE). To assess temporal consistency, we report joint velocity error (MPJVE), acceleration error, and motion jitter \cite{shimada2020physcap}, measured as the magnitude of joint jerk. We additionally report head position and orientation errors to characterise the accuracy of the estimated global motion. \\
\noindent\textbf{Baselines.}
We evaluate on four methods: EgoEgo~\cite{li2023ego}, EgoAllo~\cite{yi2025estimating}, EgoPoser~\cite{jiang2024egoposer}, and AGRoL~\cite{du2023avatars}. EgoEgo and EgoAllo estimate full-body motion using egocentric visual and head-motion observations,
whereas EgoPoser and AGRoL reconstruct body motion from sparse head and hand tracking.
All methods are evaluated using their released pretrained models. Since their required observations differ, we interpret the results under their respective input configurations. \\
\noindent\textbf{Results.}
\cref{tab:egomocap_reconstruction} presents the quantitative
results. EgoAllo achieves the lowest MPJPE (80.8 mm),
whereas AGRoL obtains the lowest world-frame joint error
(95.4 mm) and sequence-aligned error (64.9 mm).
EgoAllo additionally achieves the lowest joint velocity
error (14.5 cm/s), acceleration error (214.5 cm/s$^2$),
and motion jitter (72.06 m/s$^3$). These results reveal a discrepancy between local pose accuracy and global motion reconstruction: methods that recover accurate body articulation do not necessarily preserve the corresponding motion in world coordinates. Moreover, favourable joint-position and temporal errors alone do not establish accurate foot placement or physically plausible locomotion.
This motivates our dedicated evaluation of foot contact,
foot placement, and lower-body motion plausibility
in Sec.~\ref{sec:foot_evaluation}.

\subsection{Foot Contact and Placement Estimation}
\label{sec:foot_evaluation}

\noindent\textbf{Task.} We further evaluate lower-body reconstruction through
foot contact, placement accuracy, and kinematic plausibility for the \cref{sec:ego_reconstruction} baselines. Contact is estimated from reconstructed foot joints using a shared height- and velocity-based rule, with thresholds calibrated on the training participants. \\
\noindent\textbf{Metrics.} We evaluate foot contact with the F1 score against contact labels derived from sole markers, following prior contact evaluation protocols~\cite{rempe2020contact}. Since the baselines do not predict contact explicitly, we derive it from their foot joints using a single height and speed rule, calibrated once on ground-truth bodies of the training participants and shared by all methods. Following~\cite{rempe2021humor,yuan2023physdiff,yi2025estimating}, we measure foot skating as the horizontal speed of the foot during reference stance (\textit{Skate}), and ground penetration as the mean depth of the lowest foot joint below the floor~\cite{shimada2020physcap,yuan2023physdiff}. We further introduce two footstep-level measures: the position error of each planted foot relative to the ground-truth body (\textit{Footprint}), and the drift-free step displacement error, expressed in the frame of the preceding footprint (\textit{Step}). \\
\noindent\textbf{Results.}
Tab.~\ref{tab:feet} reports results on 543 test sessions ($\approx$18\,h), covering approximately 50k stance intervals. EgoAllo performs best across all metrics, achieving 90.5\% contact F1, 85.1\,mm footprint error, 154.1\,mm step displacement error, and 54.8\,mm/s foot skating. Despite this, substantial foot-placement errors persist. Even reference SMPL-X motion achieves only 93.6\% contact F1 against sole-marker labels, highlighting the limitations of kinematics-derived contact estimation and the need for explicit lower-body evaluation.

\subsection{Human-to-Humanoid Motion Retargeting}
\label{sec:retargeting}

\noindent\textbf{Task.} We evaluate the transfer of human motion to a 29-DoF Unitree G1  \cite{unitree_g1}, assessing both kinematic fidelity and geometric feasibility. RoboMoCap provides MoCap-derived SMPL-X motion and sole-marker-based foot-contact annotations, enabling evaluation beyond pose similarity. We retarget 543 test sessions ($\approx$18\,h, 30\,Hz), using identical input sequences and a shared robot collision model for all methods. \\
\noindent\textbf{Metrics.}
\textit{Kinematic fidelity} (Tab.~\ref{tab:egomocap_retarget_kinematic})
is evaluated against height-scaled SMPL-X motion using 13 corresponding body segments. We report mean per-body position error with a constant horizontal offset removed per sequence ($E_{\mathrm{g\text{-}mpbpe}}$) and relative to the hip midpoint ($E_{\mathrm{mpbpe}}$)~\cite{luo2023perpetual}, alongside foot and wrist position errors and geodesic orientation error. \textit{Feasibility and contact preservation}
(Tab.~\ref{tab:egomocap_retarget_feasibility}) are assessed using the robot's collision geometry. We report foot-contact F1 against sole-marker labels, foot skating (speed $>0.1$\,m/s during shared stance), ground penetration (above 5\,mm and maximum depth), and foot floating during human stance. We additionally measure near-joint-limit frames (within 0.01\,rad) and self-collision frames (penetration $>5$\,mm). A clip is considered successful if its root-relative error is below 15\,cm for at least 99\% of frames, maximum ground penetration is below 5\,cm, and self-collision
occurs in at most 5\% of frames. \\
\noindent\textbf{Baselines.}
We compare three retargeting approaches: GMR~\cite{araujo2026retargeting}, which solves
inverse kinematics on scaled human targets; PHC~\cite{luo2023perpetual}, whose shape-fitting and motion-optimisation procedure we adapt to G1; and OmniRetarget~\cite{yang2026omniretarget}, which incorporates contact and kinematic constraints. \\
\noindent\textbf{Results.} OmniRetarget achieves the highest kinematic fidelity,
reducing global body-position error from 90.7\,mm (GMR) to 34.8\,mm, with approximately fivefold lower wrist error. It also achieves the best contact preservation (87.7\% F1), lowest skating rate (2.2\%), and minimal ground penetration (0.2\,cm maximum). PHC achieves low root-relative error (64.9\,mm) but substantially higher global error (173.9\,mm) and ground penetration (up to 37.1\,cm). GMR exhibits ground penetration in 72.5\% of frames, yielding only 7.4\% clip success. Despite its superior motion fidelity, OmniRetarget achieves 53.6\% clip success, compared with 79.6\% for PHC, which exhibits substantially fewer self-collisions (1.1\%). All methods approach joint limits in over 70\% of frames, while OmniRetarget and GMR exhibit approximately 10\% self-collision rates. These findings reveal a trade-off between motion fidelity and geometric feasibility, highlighting the challenges of retargeting diverse whole-body movements to humanoid robots.


\section{Conclusion}
\label{sec:conclusion}

We introduced RoboMoCap, a large-scale multimodal dataset that bridges egocentric perception and high-fidelity whole-body motion capture, with a particular emphasis on locomotion and loco-manipulation. By combining synchronised wearable sensing with marker-based optical motion capture,
RoboMoCap provides accurate motion supervision for lower-body movements and foot placements that are often unobservable from egocentric views. Its structured collection protocol captures diverse locomotion patterns, whole-body
dynamics, and object interactions, enabling systematic investigation of how humans coordinate locomotion and manipulation. Our benchmarks provide reference evaluations for egocentric full-body motion reconstruction, foot contact and placement estimation, and human-to-humanoid retargeting. By bridging egocentric perception and accurate whole-body motion, RoboMoCap opens new opportunities for learning coordinated locomotion and manipulation from human demonstrations.

\noindent \textbf{Limitations.} Despite its scale and motion diversity, RoboMoCap is collected within an instrumented optical motion capture
environment, limiting the range of environments and terrain conditions compared with unconstrained real-world settings. Although marker-based
tracking provides precise motion measurements, occasional marker occlusions and body-model fitting errors may affect the resulting annotations. Furthermore, the dataset covers a finite set of participants, objects, and interaction scenarios, and does not represent the full variability
of human locomotion and manipulation. Finally, differences in human and humanoid morphology, kinematics, and physical capabilities mean that recorded motions cannot always be directly transferred to robotic platforms without
retargeting and control adaptations. Extending RoboMoCap to more diverse environments and validating its utility through downstream humanoid execution are promising directions for future work.


{
    \small
    \bibliographystyle{ieeenat_fullname}
    \bibliography{main}
}


\end{document}